\documentclass[sigconf]{acmart}

\setcopyright{none}
\renewcommand\footnotetextcopyrightpermission[1]{}
\acmConference[]{}{}{}

\AtBeginDocument{%
  }

\begin{document}

\title{Maat: The Agentic Legal Research Assistant for Competition Protection}

\author{Basant Mounir}
\authornote{This research paper has been prepared by its authors and does not represent the views, opinions, or official position of the Central Bank of Egypt, as their employees. The contents are provided for informational purposes only and shall not create, give rise to, or be construed as creating any legal obligation, liability, or responsibility on the part of the Central Bank of Egypt and it assumes no responsibility or liability for the content, conclusions, analyses, or recommendations contained in this paper.}
\email{basant.mounir@gmail.com}
\orcid{0009-0007-4188-8076}
\affiliation{%
  \city{Cairo}
  \country{Egypt}
}

\author{Farida Madkour}
\authornotemark[1]
\email{faridamadkour@aucegypt.edu}
\orcid{0009-0004-0646-7637}
\affiliation{%
  \city{Cairo}
  \country{Egypt}
}

\author{Amira Abdelaziz}
\authornotemark[1]
\email{amira@mathematik.uni-marburg.de}
\orcid{0009-0002-6636-9188}
\affiliation{%
  \city{Cairo}
  \country{Egypt}
}

\author{Asmaa Sami}
\authornotemark[1]
\email{asmaasami2011@gmail.com}
\orcid{0009-0009-7423-1083}
\affiliation{%
  \city{Cairo}
  \country{Egypt}
}

\begin{abstract}
Competition law experts conducting legal research must review extensive volumes of cases, decisions, and judicial reports to identify precedents and assess key elements in competition and merger cases. Although general research assistants such as Claude and ChatGPT and legal assistants such as SaulLM-7B and LegalGPT are increasingly used to assist legal research, they remain inadequate for competition law analysis: they lack specialized domain expertise, provide insufficient official citations, or hallucinate competition law cases. We propose Maat, a ReAct agent that orchestrates tools corresponding to different tasks of the research process. Designed iteratively with competition law experts, Maat grounds cases and findings in official sources using RAG for reliability, provides rich in-line citations, falls back to web search when database coverage is insufficient, and prompts the user for clarification when queries are ambiguous. Maat significantly outperforms all baseline assistants on case-specific tasks and performs within range of the top baseline on theoretical question tasks. The dataset used is available on GitHub\footnote{https://github.com/baahmed/maat-dataset}.
\end{abstract}



\maketitle

\section{Introduction}
Competition protection promotes fair competition for a healthy economic environment \cite{oecd2025}. Experts monitor markets, receive complaints, and initiate proceedings, requiring them to research cases across jurisdictions \cite{ec_procedures2024, ginsburg_eicke2023}, analyze hundreds of pages of case documents, and expand on competition law concepts \cite{oecd2024proof} using sources such as the EU Competition Commission, Concurrences \cite{concurrences2025}, and OECD reports \cite{oecd2025}. Due to the time-consuming nature of this process, experts increasingly turn to AI assistants for reasoning, summarization, and question-answering \cite{wei2022chain, deroy2024applicability, louis2024interpretable}. These assistants include general research assistants such as ChatGPT \cite{openai2026gpt55} and Claude Sonnet \cite{anthropic2026sonnet46}, and multilingual legal assistants such as SaulLM-7B \cite{colombo2024saullm} and LegalGPT \cite{legalgpt2025}. However, general research assistants are inadequate for competition law research: they hallucinate cases and case information \cite{huang2025survey, el2024factuality}, possess shallow domain knowledge \cite{ariai2025natural, el2024factuality}, and lack page-level in-line citations for verification \cite{gao2023enabling}. Legal assistants provide responses that are not grounded in official sources, omit granular citations, and perform poorly on questions requiring complex reasoning about legal principles \cite{ujwal2024reasoning, sadowski2025verifiable}.

The literature has explored more specialized systems. Much of it focuses on case retrieval or question answering in isolation \cite{kulkarni2025legal, liu2023leveraging, zhang2023diverse, locke2022case, wahidur2025legal}, while systems that integrate multiple tasks are deficient in fallback mechanisms and multi-turn clarification \cite{wang2025mars, yao2025elevating}. Crucially, none of the reviewed systems is grounded in a dedicated competition law case database, leaving the domain's cross-jurisdictional retrieval and source-level verification requirements unaddressed. To our knowledge, no prior academic work has developed and openly evaluated a purpose-built AI research assistant grounded in an official, dedicated competition law case database with cross-jurisdictional retrieval, structured tool routing, and expert evaluation.

To address these gaps, we introduce Maat, a multi-turn agentic system for competition law research, named after the ancient Egyptian goddess of law, order, and justice \cite{maat_wiki2025}. Designed iteratively with competition law experts, Maat has a structured tool-routing architecture where a ReAct-prompted agent \cite{yao2022react} dynamically selects among specialized tools based on query type and database coverage. Our contributions are as follows:
\begin{itemize}
\item a ReAct agentic system with a structured tool-routing architecture with explicit fallback logic when database coverage is insufficient and human-in-the-loop to clarify vague queries,
\item an interface to interact with the agentic research system conversationally, get page-level in-line citations from official sources, and inspect which tools are executed at each reasoning step,
\item a competition cases dataset consolidating EU regional cases and German national cases, enriched by metadata extracted by a Large Language Model (LLM) to power semantic case search,
\item a blind comparative expert evaluation demonstrating that Maat significantly outperforms all assistants under consideration in case-specific questions and performs within range of the top baseline in theoretical questions.
\end{itemize}

\section{Data}
The dataset is constructed from two sources: the EU Competition Commission and the Bundeskartellamt, chosen for their jurisdictional and financial market relevance to Egypt.

\subsection{European Union Regional Cases}
For regional cases across the EU, there are two separate API endpoints for antitrust cases and merger cases \cite{dgcomp2025cases}. Relevant metadata fields are selected, including case ID, description, violation, companies, sector, decision text link, date of release, and language of the text.

The dataset was then cleaned. Cases with empty links to the decision text were removed due to inaccessibility to the RAG pipeline. To normalize the violations, older articles were mapped to their newer counterparts. For example, for cartel cases, Article 85 EEC was mapped to Article 101 TFEU. The sectors of four cases were missing, which were filled manually. A comma-separated list of companies was extracted per case to be readable in Python. Only cases in English were kept. Other columns have passed data quality checks for missing values and normalized formats. 

\subsection{German National Cases}
For the Bundeskartellamt cases listed in \cite{bundeskartellamt2024entscheidungen}, there was no direct API or dataset source; the dataset had to be created. The strategy was to compile decision text links of Bundeskartellamt cases and extract metadata from the link structure and the decision texts using an LLM. The links of the case decision texts were studied for patterns. Cases from the Bundeskartellamt start with the prefix \url{bundeskartellamt.de/SharedDocs/Entscheidung/} and were retrieved using the Wayback Machine API \cite{internetarchive2013waybackapi}. 

The same aforementioned case metadata was extracted. From the link structure, the case ID, violation, decision release year, and language could be deduced. The remainder of the fields, specifically the case title, the sector, and the list of companies involved, were extracted by prompting the OpenAI \texttt{gpt-4o-mini} model via few-shot examples to guide the model. Since Germany is part of the EU, it also builds its economic classification system for market definition based on NACE \cite{beuter2025approaches}. To verify LLM extraction quality, a stratified random sample of 34 cases was manually reviewed; extraction accuracy was 85\% for case titles, 97\% for sectors, and 84\% for company lists.

\subsection{Combined Case Dataset}
The final combined datasets of both the regional EU cases and the national German cases consist of 1609 cases. It contains the following fields:
\begin{itemize}
  \item \texttt{case\_id}: the unique identifier of the case,
  \item \texttt{case\_title}: a description of the case,
  \item \texttt{jurisdiction}: the covered geography of the case (either EU or Germany),
  \item \texttt{violation}: the legal basis under consideration in the case,
  \item \texttt{sector}: the code and name of the NACE section that includes the sector in which the violation was committed,
  \item \texttt{companies}: a list of companies involved in the case,
  \item \texttt{pdf\_url}: the link to the case decision text,
  \item \texttt{language}: the language in which the case decision text is written,
  \item \texttt{decision\_date}: the date on which the case decision text was released.
\end{itemize}

\subsubsection{Case Decision Documents Ingestion Pipeline}
To prepare the case decision documents for querying in the RAG pipeline, the cases dataset must be indexed appropriately. \texttt{LlamaIndex} \cite{llamaindex_rag} is the framework used to process the documents for Maat. The \texttt{PDFReader} loads a decision document from its respective link into a \texttt{Document} object. The \texttt{SentenceSplitter} chunks the document into sets of \texttt{Nodes}, each consisting of approximately 1024 tokens, with 20 overlap tokens \cite{llamaindex_loading}. Each \texttt{Node} contains document-related metadata, such as the link of the document and the page from which the \texttt{Node} comes. This document-level metadata is crucial for in-line citations. The chunks are embedded using the OpenAI \\ \texttt{text-embedding-ada-002} model \cite{llamaindex_embeddings} and stored in the \texttt{Qdrant} vector database.

\begin{figure*}[!h]
    \includegraphics[width=0.99\textwidth]{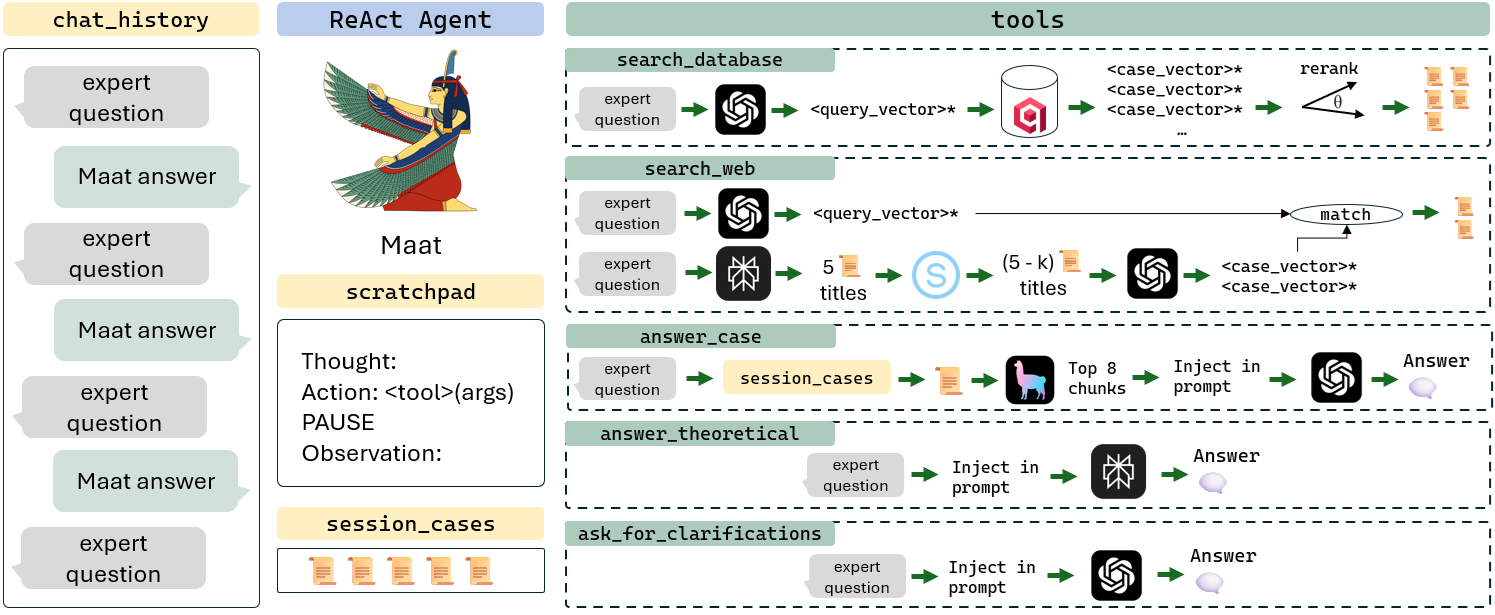}
    \hfill
    \begin{minipage}[b]{0.99\textwidth}
        \caption{Maat System Architecture. The Maat Goddess image is obtained from \cite{maat_pic_wikipedia2025}.}
        \label{fig:maat}
    \end{minipage}
\end{figure*}

\section{Architecture}

\subsection{Agent Design}
At the heart of Maat is a \texttt{ReAct} agent that mimics the expert research workflow through a thought-action-observation loop \cite{yao2022react}. The system prompt of Maat, developed by meta-prompting, describes the loop structure, the tools available, the routing rules, and the constraints.

\subsubsection{Tools}
The following tools were defined based on the observed expert research workflow: \texttt{database\_search}, \texttt{web\_search}, \texttt{answer\_case}, \texttt{answer\_theoretical}, and \texttt{ask\_clarification}. 

These tools can be chained together to completely answer the user's question. For example, if the user asks \textit{"What was the market definition in the case AT.39398?"} and the user has never referenced that case before, a \texttt{database\_search} operation is executed first. If the case was not found in the database, then this tool fails over to the \texttt{web\_search} tool. If no such case was found on the web, the agent concludes that no such case exists. However, if the case was found, then it is fetched into the memory layer of the agent. Next, the tool \texttt{answer\_case} is called to answer the question asked by the user about that case. Finally, the answer is returned. At every turn, the agent reasons about user intent, selects a tool, and observes the result.

\subsubsection{Memory Layer}
The agent uses both in-session and external memory to infuse the appropriate context in its prompts. In in-session memory, the agent records its ReAct loops in the \texttt{scratchpad} to keep track of its line of thought when making new decisions. The chat history is also stored to help the agent understand its interactions with the users. Finally, the cases retrieved so far along with their metadata are stored in \texttt{session\_cases}. \texttt{session\_cases} are crucial to answer questions such as \textit{"What was the violation in the first case?"}; the first case recorded is fetched and the question about it is answered using RAG \cite{lewis2020retrieval} as explained in \texttt{answer\_case}. The main external memory component used is the vector database which holds most of the EU regional and German national cases.

\subsection{Tools}
\subsubsection{Database Search}
If a search intention is detected in the query, the agent first tries to find matching cases in the database. The query itself is translated into a query vector composed of six dimensions that are a subset of the fields in the database: \texttt{case\_id}, \texttt{case\_title}, \texttt{jurisdiction}, \texttt{violation}, \texttt{sector}, and \texttt{companies}. 

The extraction of those dimensions from the query is performed by prompting the OpenAI \texttt{gpt-4o-mini} model, which is a recommended efficient model in terms of both cost and latency for structured data extraction \cite{openai2024gpt4omini}. The prompt includes few-shot examples per field; schema-constrained dimensions (\texttt{jurisdiction}, \texttt{violation}, \texttt{sector}) are validated against predefined values.

Since all six dimensions of the query vector are columns in the cases database, the query vector is used as a filter to match relevant cases to the query itself. The filtering of case ID is straightforward, since matches should have the exact specified ID in the query vector as the case. Exact matches are also found for the jurisdiction, violation, and sector fields, whose values come from a predefined schema. Companies are matched by checking if the companies list from the query vector is a subset of the companies list of the case under consideration. Lastly, case titles are matched using cosine similarities. Case titles of both the query vector and the case under consideration are embedded using the OpenAI \texttt{text-embedding-3-small} model, whose family has been used by other AI legal systems in \cite{text_embedding_3_1} and \cite{text_embedding_3_2}. If the cosine similarity of the embeddings exceeds the threshold of 0.85, the titles are considered a match. The threshold was tuned by testing on the case question dataset described in the evaluation section. If there are more than five cases in the result set, the cosine similarity of the query embedding and the case embeddings are computed, and only the top five cases are kept.

\subsubsection{Web Search}
Sometimes, cases are not captured because they have been released before the database is refreshed or because there are no snapshots on the WayBack Machine yet \cite{internetarchive2013waybackapi}. In this case, the agent searches the web for fresh cases. 

As with \texttt{database\_search}, the user question first gets translated to the query vector comprising the same six dimensions, which will be used to filter web results. To perform the search, Perplexity \texttt{Sonar} \cite{perplexity2025sonar} is prompted to answer the user question by providing up to five recent cases using its online search feature. Because assistants may hallucinate \cite{huang2025survey} and unofficial sources used by assistants may misinterpret cases, the web is searched for each title in the result set retrieved by Perplexity to find matching official sources from the European Competition Commission or the Bundeskartellamt, depending on the jurisdiction of the query. The Serper Google Search API \cite{serper2025} is used to search the web to gather official links for each of the retrieved titles. Titles with no official sources found are assumed hallucinated and are removed from the result set. To ensure the remaining case titles are indeed relevant to the initial query, the descriptions of the official sources for each title are translated into a six-dimensional query vector for post-search filtering. The query vector of each case title is checked against the query vector of the user question. If there is a mismatch, the corresponding case title is removed from the result set. In the end, the titles remaining are of existing cases whose metadata is a match for the original user query.

\begin{table*}
  \caption{Average Expert Answer Ratings (out of 5)}
  \label{tab:expert-ratings}
  \begin{tabular}{lccccc}
    \toprule
    & & \multicolumn{2}{c}{\textbf{General Research LLMs}} & \multicolumn{2}{c}{\textbf{Legal LLMs}} \\
    \cmidrule(lr){3-4} \cmidrule(lr){5-6}
    Task Category & Maat & Claude Sonnet 4.6 & GPT-5.5 & LegalGPT & SaulLM-7B \\
    \midrule
    Theoretical   & 2.9 & 3.5 & 2.4 & 2.2 & 1.1 \\
    Case-specific & 4.6 & 2.4 & 2.0 & 2.1 & 0.7 \\
    \bottomrule
  \end{tabular}
  \vspace{2pt}
  \\[2pt]
\end{table*}

\subsubsection{Answer Case-Specific Questions}
When a user asks a question about a specific case, the agent first checks if the case in question exists in the \texttt{session\_cases} list. If not, a search operation is invoked before continuing to retrieve the case information. The chunks of the case in question are loaded in \texttt{active\_case} from the database. Then, the RAG \cite{lewis2020retrieval} pipeline is invoked. The case chunks are ranked according to the cosine similarities of each chunk and the embedded user question. Based on experimentation using the question bank described in the next section, the number of top chunks to be retrieved was set to ${8}$. The texts, pages, and links of the top ${8}$ chunks are injected into a meta-prompted question-answering prompt alongside the question and chat history to generate the final answer with in-line citations.

\subsubsection{Answer Theoretical Questions}
When a user asks a general question about the theory of competition law, such as \textit{"What are the forms of abuse of dominance?"}, \texttt{answer\_theoretical} is invoked. A prompt was engineered iteratively with competition law experts to construct the knowledge base around key concepts and terminologies of competition law. Perplexity \texttt{Sonar} was the base LLM of choice due to its deep research capabilities \cite{perplexity2025sonar}. The question is injected into this prompt to guide Perplexity's reasoning depth and direction. Furthermore, the domain of search was restricted to a set of official and reliable sources only provided by the experts. At the end of the response, based on the chat history, Maat also recommends questions for deeper research in its role as a brainstorming assistant.

\subsubsection{Ask for Clarification}
If the question is unclear, the scratchpad, chat history, and question are analyzed to assess the information gap that needs to be addressed before the agent answers the question. For example, if the user asks \textit{"What was the market definition of the case?"} and there are multiple retrieved cases by a search operation, then the agent will first ask the user to clarify which case is the question about. The \texttt{human-in-the-loop} then provides enough clarity for the agent to answer the question.

\subsection{User Interface}
Maat is a web application accessible via standard browsers without additional installation. The Flask library is used to design an interactive user experience with a dedicated user input box and submit button to enter questions. As the user question is passed to the agent in the backend, the user can see a progress indicator, \textit{"The assistant is thinking..."} that adjusts to show which tools the agent calls during the question-answering process. The user can track the conversation in the scrollable chat pane and explore in-line citations for each part of the answer, allowing for a smooth answer verification experience.

\section{Evaluation}
A blind comparative evaluation was conducted in which the performance of Maat was compared against state-of-the-art general research assistants, ChatGPT (GPT-5.5) ~\cite{openai2026gpt55} and Claude Sonnet 4.6 \cite{anthropic2026sonnet46}, and specialized multilingual legal research assistants, SaulLM-7B~\cite{colombo2024saullm} and LegalGPT~\cite{legalgpt2025}. The experts prepared two question banks: fifty theoretical questions (e.g., \textit{"Explain price-fixing"}) and twenty case-specific questions (e.g., \textit{"List abuse of dominance cases in the financial sector"}). Seven questions from the theoretical test bank and eight questions from the case-specific test bank formed the test set; the remainder were used for prompt engineering and hyperparameter tuning. Responses were scored from 1 (poor) to 5 (excellent) based on five criteria, each worth one point: coherence of legal logic, usage of correct legal taxonomy, illustration of concepts, quality of references, and appropriate inclusion of in-line citations. Nine competition law experts contributed to the evaluation process.

\subsection{Results}
Table~\ref{tab:expert-ratings} reports average expert ratings per task type. Maat substantially outperformed all baselines on case-specific questions and performed within range of Claude Sonnet 4.6, the strongest baseline, on theoretical questions; a Friedman test confirmed these differences are statistically significant (${p < 0.05}$). Maat and Claude Sonnet 4.6 both demonstrated structured legal reasoning and illustrative examples, though both lacked citation granularity and source diversity, with Claude Sonnet 4.6 performing marginally better on these criteria. On case-specific questions, legal-domain assistants performed poorly, with experts observing hallucinated cases, unofficial sources, missing in-line citations, and weak competition law reasoning. This reflects both slower research cycles in legal AI and the lack of competition law specialization of these models. General research assistants retrieved more relevant cases but shared the same citation and reasoning limitations. Maat addresses these shortcomings by design: it retrieves case documents directly from official sources, generates answers via RAG with page-level in-line citations, and applies structured prompt engineering for principled competition law reasoning.

\section{Conclusion}
We introduced Maat, a ReAct agent for competition law research that integrates database retrieval, web fallback, case and theoretical question-answering, and human-in-the-loop clarification. Grounded in Bundeskartellamt and EU Commission cases with in-line citations for verifiability, Maat performed within range of the top baseline in theoretical question tasks and outperformed all baselines across case-specific tasks. Future work will extend jurisdictional coverage and enhance theoretical source diversity and citation granularity.

\begin{acks}
The authors thank the EU Competition Commission and the Bundeskartellamt for making the cases publicly accessible, and Ahmed Abdelaal, Abdelrahman Abdelhalim, Ahmed Kamel, and Waad Hegazy for paper review and involvement throughout the development of Maat.
\end{acks}

\section*{GenAI Usage Disclosure}
The authors utilized Claude and ChatGPT to debug code written for Maat. Furthermore, the authors used the aforementioned GenAI tools for grammar checking, minor paper structuring, and draft revisions. All GenAI responses were checked and edited for accuracy and appropriateness. GenAI was not used to contribute to the intellectual content. The authors claim full responsibility for all contributions and results of this publication.

\bibliographystyle{ACM-Reference-Format}
\bibliography{sample-base}

@String{Computing = "Computing" }

@String{Springer = "Springer-Verlag" }

@article{deroy2024applicability,
  title={Applicability of large language models and generative models for legal case judgement summarization},
  author={Deroy, Aniket and Ghosh, Kripabandhu and Ghosh, Saptarshi},
  journal={arXiv preprint arXiv:2407.12848},
  year={2024}
}

@misc{text_embedding_3_1,
      title={All for law and law for all: Adaptive RAG Pipeline for Legal Research}, 
      author={Figarri Keisha and Prince Singh and Pallavi and Dion Fernandes and Aravindh Manivannan and Ilham Wicaksono and Faisal Ahmad and Wiem Ben Rim},
      year={2025},
      eprint={2508.13107},
      archivePrefix={arXiv},
      primaryClass={cs.CL},
      url={https://arxiv.org/abs/2508.13107}, 
}

@misc{text_embedding_3_2,
      title={LegalBench-RAG: A Benchmark for Retrieval-Augmented Generation in the Legal Domain}, 
      author={Nicholas Pipitone and Ghita Houir Alami},
      year={2024},
      eprint={2408.10343},
      archivePrefix={arXiv},
      primaryClass={cs.AI},
      url={https://arxiv.org/abs/2408.10343}, 
}

@misc{openai2024gpt4omini,
  author       = {{OpenAI}},
  title        = {{GPT-4o mini: Advancing Cost-Efficient Intelligence}},
  year         = {2024},
  month        = jul,
  howpublished = {\url{https://openai.com/index/gpt-4o-mini-advancing-cost-efficient-intelligence/}},
  }

@misc{dgcomp2025cases,
  author       = {{Directorate-General for Competition, European Commission}},
  title        = {{EU Competition Case Search}},
  year         = {2026},
  howpublished = {\url{https://competition-cases.ec.europa.eu/search}},
  note         = {Official European Commission database for antitrust, cartel, merger, and state aid cases distributed in JSON format. License: European Commission Reuse Notice (Dec.~2011/833/OJ).}
}

@misc{bundeskartellamt2024entscheidungen,
  author       = {{Bundeskartellamt}},
  title        = {{Entscheidungen [Decisions]}},
  year         = {2024},
  howpublished = {\url{https://www.bundeskartellamt.de/SharedDocs/Entscheidung/}},
  note         = {Official decision database of the German Federal Cartel Office, 
                  published pursuant to \S~5 UrhG.}
}

@misc{internetarchive2013waybackapi,
  author       = {{Internet Archive}},
  title        = {{Wayback Machine APIs}},
  howpublished = {\url{https://archive.org/help/wayback_api.php}},
  note         = {Documents the Wayback Availability JSON API and CDX Server API.}
}

@Inbook{beuter2025approaches,
author="Beuter, Felix
and Gussenbauer, Johannes
and Minther, Elias
and Szabo, Viktoria
and Wegner, Susanne",
editor="Dumpert, Florian",
title="Approaches to Automated NACE Coding of German Business Activity Descriptions",
bookTitle="Foundations and Advances of Machine Learning in Official Statistics",
year="2025",
publisher="Springer Nature Switzerland",
address="Cham",
pages="179--211",
isbn="978-3-032-10004-7",
doi="10.1007/978-3-032-10004-7_10",
url="https://doi.org/10.1007/978-3-032-10004-7_10"
}

@misc{llamaindex_embeddings,
  author       = {{LlamaIndex}},
  title        = {Embeddings},
  year         = {2025},
  url          = {https://developers.llamaindex.ai/python/framework/module_guides/models/embeddings/},
  note         = {LlamaIndex Developer Documentation.}
}

@misc{llamaindex_loading,
  author       = {{LlamaIndex}},
  title        = {Loading Data (Ingestion)},
  year         = {2025},
  url          = {https://developers.llamaindex.ai/python/framework/understanding/rag/loading/},
  note         = {LlamaIndex Developer Documentation.}
}

@misc{llamaindex_rag,
  author       = {{LlamaIndex}},
  title        = {Introduction to {RAG}},
  year         = {2025},
  url          = {https://developers.llamaindex.ai/python/framework/understanding/rag/},
  note         = {LlamaIndex Developer Documentation. }
}

@article{yao2022react,
  title={React: Synergizing reasoning and acting in language models},
  author={Yao, Shunyu and Zhao, Jeffrey and Yu, Dian and Du, Nan and Shafran, Izhak and Narasimhan, Karthik and Cao, Yuan},
  journal={arXiv preprint arXiv:2210.03629},
  year={2022}
}

@article{huang2025survey,
  title={A survey on hallucination in large language models: Principles, taxonomy, challenges, and open questions},
  author={Huang, Lei and Yu, Weijiang and Ma, Weitao and Zhong, Weihong and Feng, Zhangyin and Wang, Haotian and Chen, Qianglong and Peng, Weihua and Feng, Xiaocheng and Qin, Bing and others},
  journal={ACM Transactions on Information Systems},
  volume={43},
  number={2},
  pages={1--55},
  year={2025},
  publisher={ACM New York, NY}
}

@misc{serper2025,
  author       = {{Serper}},
  title        = {Serper: Google Search API},
  year         = {2025},
  url          = {https://serper.dev/},
}

@article{lewis2020retrieval,
  title={Retrieval-augmented generation for knowledge-intensive nlp tasks},
  author={Lewis, Patrick and Perez, Ethan and Piktus, Aleksandra and Petroni, Fabio and Karpukhin, Vladimir and Goyal, Naman and K{\"u}ttler, Heinrich and Lewis, Mike and Yih, Wen-tau and Rockt{\"a}schel, Tim and others},
  journal={Advances in neural information processing systems},
  volume={33},
  pages={9459--9474},
  year={2020}
}

@misc{maat_pic_wikipedia2025,
  author       = {{Eternal Space}},
  title        = {Maat (Goddess)},
  url          = {https://commons.wikimedia.org/wiki/File:Maat_(Goddess).png},
  note         = {Licensed under CC BY-SA 4.0}
}

@inproceedings{kulkarni2025legal,
  title={Legal Case Search: An AI-Powered Legal Search Engine},
  author={Kulkarni, Radhika V and Agrawal, Avish and Vimal, Aryan and Barde, Rohan and Bajaj, Raghav and Gaddi, Khursheed},
  booktitle={International Conference on ICT for Sustainable Development},
  pages={354--363},
  year={2025},
  organization={Springer}
}

@inproceedings{liu2023leveraging,
  title={Leveraging event schema to ask clarifying questions for conversational legal case retrieval},
  author={Liu, Bulou and Hu, Yiran and Ai, Qingyao and Liu, Yiqun and Wu, Yueyue and Li, Chenliang and Shen, Weixing},
  booktitle={Proceedings of the 32nd ACM international conference on information and knowledge management},
  pages={1513--1522},
  year={2023}
}

@article{zhang2023diverse,
  title={Diverse legal case search},
  author={Zhang, Ruizhe and Ai, Qingyao and Wu, Yueyue and Ma, Yixiao and Liu, Yiqun},
  journal={arXiv preprint arXiv:2301.12504},
  year={2023}
}

@article{locke2022case,
  title={Case law retrieval: problems, methods, challenges and evaluations in the last 20 years},
  author={Locke, Daniel and Zuccon, Guido},
  journal={arXiv preprint arXiv:2202.07209},
  year={2022}
}

@article{wahidur2025legal,
  title={Legal query rag},
  author={Wahidur, Rahman SM and Kim, Sumin and Choi, Haeung and Bhatti, David S and Lee, Heung-No},
  journal={IEEE Access},
  year={2025},
  publisher={IEEE}
}

@inproceedings{yao2025elevating,
  title={Elevating legal LLM responses: harnessing trainable logical structures and semantic knowledge with legal reasoning},
  author={Yao, Rujing and Wu, Yang and Wang, Chenghao and Xiong, Jingwei and Wang, Fang and Liu, Xiaozhong},
  booktitle={Proceedings of the 2025 Conference of the Nations of the Americas Chapter of the Association for Computational Linguistics: Human Language Technologies (Volume 1: Long Papers)},
  pages={5630--5642},
  year={2025}
}

@article{ariai2025natural,
  title={Natural language processing for the legal domain: A survey of tasks, datasets, models, and challenges},
  author={Ariai, Farid and Mackenzie, Joel and Demartini, Gianluca},
  journal={ACM Computing Surveys},
  volume={58},
  number={6},
  pages={1--37},
  year={2025},
  publisher={ACM New York, NY}
}

@misc{concurrences2025,
  author       = {{Concurrences}},
  title        = {Concurrences: Competition Law Review},
  url          = {https://www.concurrences.com/en/},
}

@misc{oecd2025,
  author       = {{Organisation for Economic Co-operation and Development}},
  title        = {{OECD}},
  url          = {https://www.oecd.org/en.html},
}

@article{wei2022chain,
  title={Chain-of-thought prompting elicits reasoning in large language models},
  author={Wei, Jason and Wang, Xuezhi and Schuurmans, Dale and Bosma, Maarten and Xia, Fei and Chi, Ed and Le, Quoc V and Zhou, Denny and others},
  journal={Advances in neural information processing systems},
  volume={35},
  pages={24824--24837},
  year={2022}
}

@inproceedings{louis2024interpretable,
  title={Interpretable long-form legal question answering with retrieval-augmented large language models},
  author={Louis, Antoine and Van Dijck, Gijs and Spanakis, Gerasimos},
  booktitle={Proceedings of the AAAI conference on artificial intelligence},
  volume={38},
  number={20},
  pages={22266--22275},
  year={2024}
}

@inproceedings{gao2023enabling,
  title={Enabling large language models to generate text with citations},
  author={Gao, Tianyu and Yen, Howard and Yu, Jiatong and Chen, Danqi},
  booktitle={Proceedings of the 2023 Conference on Empirical Methods in Natural Language Processing},
  pages={6465--6488},
  year={2023}
}

@misc{ec_procedures2024,
  author       = {{European Commission}},
  title        = {Antitrust and Cartels: Procedures},
  url          = {https://competition-policy.ec.europa.eu/antitrust-and-cartels/procedures_en}
}

@book{ginsburg_eicke2023,
  editor       = {Ginsburg, Douglas H. and Eicke, Tim},
  title        = {Judicial Review of Competition Cases},
  publisher    = {Concurrences},
  year         = {2023},
  note         = {Multi-jurisdictional comparative study}
}

@techreport{oecd2024proof,
  author       = {{OECD}},
  title        = {The Standard and Burden of Proof in Competition Law Cases},
  institution  = {OECD Competition Committee},
  year         = {2024},
  url          = {https://doi.org/10.1787/0199f63f-en},
}

@misc{maat_wiki2025,
  author       = {{Wikipedia contributors}},
  title        = {Maat},
  url          = {https://en.wikipedia.org/wiki/Maat},
}

@misc{legalgpt2025,
  author       = {{LexiAI}},
  title        = {Legal {GPT}},
  howpublished = {\url{https://chatgpt.com/g/g-jxqQ0lepc-legal-gpt}},
}

@inproceedings{el2024factuality,
  title={The factuality of large language models in the legal domain},
  author={El Hamdani, Rajaa and Bonald, Thomas and Malliaros, Fragkiskos D and Holzenberger, Nils and Suchanek, Fabian},
  booktitle={Proceedings of the 33rd ACM International Conference on Information and Knowledge Management},
  pages={3741--3746},
  year={2024}
}

@inproceedings{ujwal2024reasoning,
  title={" Reasoning before Responding": Towards Legal Long-form Question Answering with Interpretability},
  author={Ujwal, Utkarsh and Surampudi, Sai Sri Harsha and Mitra, Sayantan and Saha, Tulika},
  booktitle={Proceedings of the 33rd ACM International Conference on Information and Knowledge Management},
  pages={4922--4930},
  year={2024}
}

@inproceedings{sadowski2025verifiable,
  title={On verifiable legal reasoning: A multi-agent framework with formalized knowledge representations},
  author={Sadowski, Albert and Chudziak, Jaroslaw A},
  booktitle={Proceedings of the 34th ACM International Conference on Information and Knowledge Management},
  pages={2535--2545},
  year={2025}
}

@article{wang2025mars,
  title={L-MARS: Legal multi-agent workflow with orchestrated reasoning and agentic search},
  author={Wang, Ziqi and Yuan, Boqin},
  journal={arXiv preprint arXiv:2509.00761},
  year={2025}
}

@misc{perplexity2025sonar,
  author       = {{Perplexity AI}},
  title        = {Meet New {Sonar}},
  year         = {2025},
  howpublished = {Perplexity Blog},
  url          = {https://www.perplexity.ai/hub/blog/meet-new-sonar}
}

@techreport{anthropic2026sonnet46,
  title        = {{Claude Sonnet 4.6 System Card}},
  author       = {{Anthropic}},
  year         = {2026},
  month        = {February},
  url          = {https://www.anthropic.com/claude-sonnet-4-6-system-card},
}

@techreport{openai2026gpt55,
  title        = {{GPT-5.5 System Card}},
  author       = {{OpenAI}},
  year         = {2026},
  month        = {April},
  institution  = {OpenAI},
  url          = {https://openai.com/index/gpt-5-5-system-card/},
  note         = {Accessed: May 2026}
}

@article{colombo2024saullm,
  title={Saullm-7b: A pioneering large language model for law},
  author={Colombo, Pierre and Pires, Telmo Pessoa and Boudiaf, Malik and Culver, Dominic and Melo, Rui and Corro, Caio and Martins, Andre FT and Esposito, Fabrizio and Raposo, Vera L{\'u}cia and Morgado, Sofia and others},
  journal={arXiv preprint arXiv:2403.03883},
  year={2024}
}

\end{document}